\documentclass[runningheads]{llncs}
\usepackage{graphicx}
\usepackage{textcomp}
\usepackage{multirow}
\usepackage{multicol}
\usepackage{makecell}
\usepackage{enumitem}
\usepackage{todonotes}
\usepackage{orcidlink}

\makeatletter
\newcommand{\printfnsymbol}[1]{%
  \textsuperscript{\@fnsymbol{#1}}%
}
\makeatother

\begin{document}

%%
%% The "title" command has an optional parameter,
%% allowing the author to define a "short title" to be used in page headers.
\title{BOTracle: A framework for Discriminating Bots and Humans}

\author{Jan Kadel\inst{1} \and
August See\inst{2}\orcidlink{0009-0003-9588-7096} \and
Ritwik Sinha\inst{3}\orcidlink{0000-0002-0344-1368} \and
Mathias Fischer\inst{2}\orcidlink{0000-0002-6254-8288}}

\authorrunning{Kadel et al.}
% First names are abbreviated in the running head.
% If there are more than two authors, 'et al.' is used.
%
\institute{Adobe Inc. \email{\{kadel,risinha\}@adobe.com} \and
Universität Hamburg, Germany \email{\{richard.august.see,mathias.fischer\}@uni-hamburg.de}}

\maketitle

%%
%% The abstract is a short summary of the work to be presented in the
%% article.
\begin{abstract}
Bots constitute a significant portion of Internet traffic and are a source of various issues across multiple domains. Modern bots often become indistinguishable from real users, as they employ similar methods to browse the web, including using real browsers.
We address the challenge of bot detection in high-traffic scenarios by analyzing three distinct detection methods. The first method operates on heuristics, allowing for rapid detection. The second method utilizes, well known, technical features, such as IP address, window size, and user agent. It serves primarily for comparison with the third method. In the third method, we rely solely on browsing behavior, omitting all static features and focusing exclusively on how clients behave on a website.
In contrast to related work, we evaluate our approaches using real-world e-commerce traffic data, comprising 40 million monthly page visits. We further compare our methods against another bot detection approach, Botcha, on the same dataset. Our performance metrics, including precision, recall, and AUC, reach 98 percent or higher, surpassing Botcha.
\end{abstract}

\section{Introduction}
The automation of work through software is increasingly common, and used to alleviate the burden of monotonous tasks, particularly in internet communications, where bots automate repetitive functions. A prime example is search engine bots indexing the internet autonomously. However, the prevalence of malicious bots is concerning, with over half of internet traffic being bot-related \cite{bot-traffic-report}. Trends indicate a rise in malicious activities by bots, including spamming, scalping, and influencing elections~\cite{web-bot-detection-sequential,bot-traffic-report,bessi2016social}.

Differentiating between bot and human traffic is challenging due to the diverse and sophisticated nature of bots, some of which can mimic complete browser environments to evade detection. The issue is compounded when detection systems incorrectly identify humans as bots or fail to detect bots, especially as bots increasingly evade measures like CAPTCHAs \cite{captcha-solver-3,captcha-solver-2,bot-detection-evasion-gan,bot-detection-evasion-gan-2}. Such errors can lead to customer dissatisfaction and security risks.
Therefore, bot detection systems should prioritize identifying bots based on characteristics impervious to common evasion techniques, focusing on passive monitoring and analyzing inherent traits of web clients, such as behavior or intent, rather than relying solely on challenge-response tests \cite{captcha-solver-3,captcha-solver-2,bot-detection-evasion-gan,bot-detection-evasion-gan-2}.

The primary contribution of this paper is the development of a multi-stage web bot detection pipeline with distinct detectors, evaluated on a real-world e-commerce dataset. In greater detail, our contributions include:

\begin{itemize}
\item We propose a layered bot detection approach suitable for business environments, given our industry focus. This approach combines multiple techniques into an efficient pipeline. It begins with heuristic methods that utilize static features, such as IP addresses, to make quick decisions, primarily to conserve computing resources. For more detailed per-request analysis, a semi-supervised Generative Adversarial Network (SGAN) is employed. Additionally, a Deep Graph Convolutional Neural Network (DGCNN) evaluates potential bots using website traversal graphs, emphasizing session-wide behavior. Notably, this approach concentrates solely on behavior, without relying on features such as IP addresses, user agents, or window sizes. Bots are generally unaware of the average behavior on a specific website, making it challenging for them to mimic it. Furthermore, even if a bot attempts to mimic typical user behavior, it loses its inherent advantages, such as tirelessness and speed.

\item Our approach and its individual components are evaluated using a proprietary, anonymized dataset from a real-world e-commerce platform\footnote{Unfortunately, we cannot release this dataset or disclose the company's name, as this may violate our business contract with them}. This platform attracts approximately 40 million monthly page visits and generates an estimated annual revenue of between 500 million and 1 billion USD. We benchmark the performance of our method against related works that have utilized the same dataset.

\item We conduct an in-depth analysis to ascertain which bot features most significantly influence the SGAN's effectiveness. Furthermore, we explore how the size of website traversal graphs impacts the performance of the DGCNN.
\end{itemize}

In this paper, we first review existing bot detection methods, assessing their strengths and weaknesses (Section \ref{sec:related-work}). We then introduce our novel approach for effective bot detection (Section \ref{sec:method}), followed by an evaluation of its performance against established methods using real-world data (Section \ref{sec:eval}). The paper concludes with a summary of our findings and their implications (Section \ref{sec:conclusion}).

\section{Related Work}
\label{sec:related-work}
Research in this field can be categorized into two main streams: active and passive approaches. Active methods, such as CAPTCHAs, present direct challenges to users. Passive methods typically involve risk assessment strategies that utilize heuristics or machine learning algorithms. This section will review a selection of prior studies, highlighting their respective advantages and limitations.

\subsection{Active Approaches}
Modern CAPTCHA systems, like hCaptcha and reCAPTCHA~\cite{recaptcha,hcaptcha}, blend biometric data, including mouse movements, with request information. Despite this integration, the specifics of their bot detection strategies, such as website-specific tailoring or the effectiveness of various elements, remain undisclosed. This opacity is likely strategic, aimed at safeguarding proprietary methods and hindering bot developers from adapting their strategies. Google, for instance, keeps the data utilized in reCAPTCHA under wraps. Research efforts, including those by Sivakorn et al. \cite{sivakorn2016m}, are dedicated to deciphering these CAPTCHA mechanisms.
Moreover, CAPTCHAs, despite their technological advances, still detract from user experience and can be time-consuming. This is true even in modern systems that employ risk assessment to reduce the frequency of CAPTCHA challenges for users~\cite{HumanityWastes5002021}.

This category also includes registration with personal data, such as using telephone numbers or presenting an official ID, are included. These methods, however, may deter users due to the additional hurdles they present \cite{expedia} and their lack of privacy friendliness.

\subsection{Passive Approaches}
Bot detection can be straightforwardly implemented by blocking IPs known for malicious activities, utilizing resources such as \textit{abuseipdb}\footnote{https://www.abuseipdb.com/}. Moreover, various methods focus on analyzing request data to detect bots~\cite{requestbotdetection,hcaptcha,recaptcha,suchacka2021efficient,li2021good,jonker2019fingerprint,HILSA}. However, there is an issue wherein a bot might abandon browser automation frameworks like Selenium or Puppeteer and opt instead for a headful Chrome instance. This approach could circumvent fingerprints that are specific to browser automation frameworks.

There are also obfuscation approaches that complicate bot creation. See et al.~\cite{see2022polymorphic} proposed polymorphic protocols that aim to generate a unique protocol for each client (user). While this approach is effective for native applications, it does not perform well for web clients and within the HTTP context.

Iliou et al. \cite{requestbotdetection} compared machine learning algorithms using attributes from previous studies, avoiding cross-website tracking or external resources like IP databases for a more replicable and privacy-friendly approach. Analyzing a year's worth of HTTP log data from MK-Lab's web server\footnote{https://mklab.iti.gr/}, they categorized bots as simple or advanced based on browser agent names and IP history. Their findings indicate that the effectiveness of attribute sets varies with the chosen machine learning method, with Random Forest and Multilayer Perceptron performing best. An ensemble classifier provided more stable results. While simple bots were easily detectable, advanced bot detection proved challenging, especially in scenarios requiring low false positive rates. 

Related research also explores biometric authentication using mouse dynamics or typing behavior\cite{see2023detecting,see2024detecting,6263955,sayed2013biometric,jorgensen2011mouse,seemouse}. See et al. \cite{see2023detecting,see2024detecting} demonstrated the feasibility of distinguishing humans from bots based on mouse dynamics and typing behavior, utilizing open-source datasets for human data and generating bot data from third-party projects. They concluded that, while differentiation is currently possible, the sophistication of methods for generating artificial human-like keystrokes and mouse movements is likely to increase.

The Botcha framework~\cite{botcha} centers on identifying malicious bots, employing Positive Unlabeled Learning (PU Learning) for data labeling. It assumes e-commerce platform clients making purchases are human, while others remain unlabeled. A machine learning classifier then categorizes these unlabeled sessions, determining human-like characteristics. Dhamnani et al. modified the traditional PU classifier to increase accuracy. Their validation used cloud provider IP addresses as negative benchmarks, with the model incorrectly classifying 2.5\% of these as human but correctly identifying 99\% of human-labeled data. The framework tagged 82\% of sessions as human, leaving 18\% unclassified.
While effective in data-scarce scenarios, using purchasing behavior as a sole human indicator is problematic, given some bots are designed for purchases \cite{scalping-bots} or content scraping \cite{content-scraping}. Additionally, Botcha's reliance on authentication mechanisms to deter bots may be ineffective against advanced bots capable of autonomous authentication. Although Botcha's adaptation of PU learning improves performance with specific datasets, this approach may limit its broader applicability. We later compare our approach to Botcha's results.

Cabri et al. \cite{web-bot-detection-sequential} analyzed web sessions from HTTP logs, using heuristics for initial classification. Sessions were identified as bots based on factors like known bot user agents and IP addresses, or as humans if matching Udger database browsers \cite{udger}. Their approach used a multi-layer perceptron (MLP) and Wald's sequential probability ratio test \cite{wald-stats} to classify sessions as Bot, Human, or None, with uncertain cases reassessed using more data. This method achieved a notable F1 score of approximately 0.96. However, the lack of external benchmarking and reliance on database labels for human-bot distinction, which could be exploited by bots altering user agents, are limitations. Yet, its innovative use of common HTTP features and reassessment of ambiguous cases demonstrate its potential effectiveness.

BotGraph~\cite{bot-graph} is a web bot detection method, analyzes client sessions through website traversal patterns. It constructs a website's sitemap, then maps session requests onto a sitemap subgraph, representing client navigation. These subgraphs are transformed into images to train a Convolutional Neural Network (CNN), classifying sessions as bot or non-bot using supervised learning. Over 30 experts manually labeled the data, focusing on behavioral and fingerprinting features. The model achieved 93.5\% accuracy on a search engine dataset, 95.7\% on a news site, and 98.4\% on a university site, with overall precision and recall around 95\%.
BotGraph effectively differentiates bot and human website usage patterns. However, it faces challenges in sessions with few requests (> 3), where both bots and humans show similar patterns \cite{bot-graph}. Despite its reliance on manual labeling, which might introduce biases \cite{issues-hand-labeling}, BotGraph's adaptability across websites and its efficient performance (adding 2-6 milliseconds of latency on GPUs) are notable strengths.

The reviewed bot detection methods, while effective to some degree, each have notable limitations. Approaches relying on user agents are vulnerable to manipulation \cite{web-bot-detection-sequential}. Botcha assumes bots do not make purchases, a premise easily challenged by bots capable of transactions \cite{botcha}. BotGraph's focus on client traversal patterns overlooks critical page metadata, leading to potential false positives \cite{bot-graph}. Additionally, CAPTCHAs, commonly used for bot detection, degrade user experience and can be bypassed \cite{captcha-solver-2,captcha-solver-3}. These issues highlight the need for more robust, non-circumventable bot detection systems with high precision and recall, which current methodologies do not fully offer.

\section{Detecting Bots}
\label{sec:method}
Our methodology integrates multiple classification strategies, including heuristic techniques, a Semi-Supervised Generative Adversarial Network (SGAN), and a Deep Graph Convolutional Neural Network (DGCNN). This integration aims to automate captcha resolution, thereby enhancing user experience by minimizing human intervention.

\begin{figure}[ht]
\centering
\includegraphics[width=1.1\linewidth]{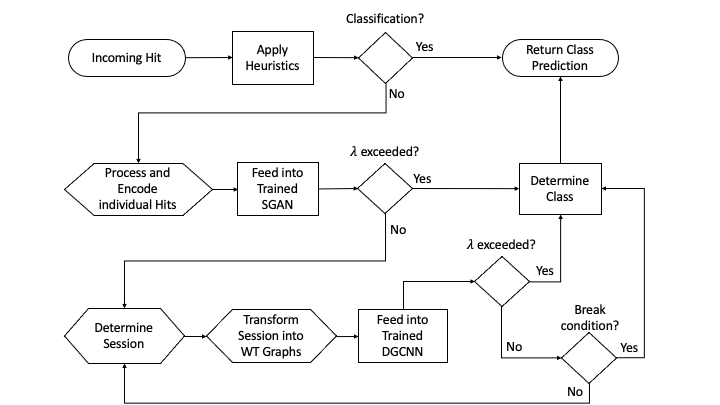}
\caption{Multi-Stage Bot Detection Pipeline Process as Flow Chart}
\label{fig:bot_detection_pipeline}
\end{figure}

Figure \ref{fig:bot_detection_pipeline} depicts our bot detection pipeline. The procedure commences with an incoming hit - a request to a web server record. Initially, rapid heuristic-based bot detection is employed. If the heuristic approach successfully classifies a hit, the process concludes by issuing a class prediction. If not, the hit undergoes processing and encoding to prepare it for the SGAN model. This preprocessing involves discarding irrelevant features and transforming the data into a one-dimensional numerical vector using flagging, integer, and one-hot encoding. Additionally, we aggregate rare categorical values into special categories to compact the input vector. Once transformed, the numerical vector is input into the SGAN, yielding a bot or human probability. This probability is evaluated against a confidence threshold $\lambda \in [0, 1]$. Probabilities below $\lambda$ lead to a rejection of the prediction due to insufficient confidence. Conversely, exceeding the threshold results in a definitive class label. Hits failing to produce a confident SGAN prediction are accumulated into client sessions, subsequently transformed into Website Traversal (WT) graphs, and analyzed by the DGCNN. Similar to the SGAN, the DGCNN produces a probability compared against $\lambda$. Surpassing this threshold finalizes the process with a class prediction, while falling short requires additional hits to augment the session before repeating the analysis.

Subsequent sections detail the pipeline components. We commence by outlining our approach to labeling in an unlabeled dataset from a website with over 50 million monthly visits, differentiating between bot and human characteristics. Generated using Adobe Analytics\textsuperscript{\textregistered} within the Adobe Experience Cloud\textsuperscript{\textregistered}, this dataset lacks pre-existing bot-related labels. Our labeling strategy is versatile, suitable for various websites. We then describe the heuristic bot detection methods, designed to identify straightforward cases like unmodified user-agent bots, thereby enriching the labeled data for SGAN and DGCNN training. Moreover, we elucidate the concept of WT graphs and their efficacy in bot detection. The machine learning models utilized in the pipeline are not detailed, as our focus was on their application rather than modifying their fundamental principles.

\subsection{Bot Feature Analysis}
\label{susec:bot-feature-analyis}
In the task of distinguishing web clients as either bots or humans, it is essential to conduct a comprehensive analysis of client attributes, identifying those most indicative of bot activity. We categorize these attributes into two main groups: behavioral and non-behavioral. Non-behavioral attributes encompass identity, and technical factors, whereas behavioral attributes are defined by patterns in navigation, interaction, and frequency of website visits.

\begin{description}
\item[Non-behavioral Attributes]:
\begin{itemize}
\item Identity Attributes: Consist of user identifiers, such as IP addresses.
\item Technical Attributes: Relate to technical aspects, like user agent and screen size.
\end{itemize}

\item[Behavioral Attributes]:
\begin{itemize}
\item Traversal Attributes: Capture user's navigation patterns on the website.
\item Interaction Attributes: Chronicle user interactions with site elements.
\item Visit Attributes: Indicate the frequency and pattern of user visits.
\end{itemize}
\end{description}

Identity attributes, such as user location, or name, are seldom used in bot detection models due to their ease of manipulation. Focus is instead placed on behavioral and technical attributes, which compel bot developers to mimic human behavior more intricately to evade detection.

Technical attributes, such as user agent, screen size, and Java applet support, provide valuable insights. Models can detect bots through unusual user agents and atypical screen dimensions. Java applet support helps identify outdated or falsified user agents. However, these can also be forged.

Traversal attributes map a client's website journey, highlighting visited pages. Patterns like exhaustive breadth-first traversal suggest bot activity, unlike humans' targeted visits. Interaction attributes chronicle client engagements, such as using promo codes and purchasing. These patterns can distinguish bot types, for instance, crawlers that avoid purchases or scalper bots aiming for bulk acquisitions. Additionally, visit attributes, including visit frequency and patterns, help detect spammer bots, especially with high visit frequency and volume.

\subsubsection{Heuristics for Detection of Obvious Bots}
\label{sususec:heuristic-bot-detection}
The primary aim of bot detection heuristics is to identify and filter out apparent bots, a process that also aids in training machine learning models. For example, a user agent self-identified as "python request" is typically indicative of a bot. Labeling all data linked to such an agent as "bot" and integrating this into the training dataset enables the machine learning model to discern patterns, like bots making requests at regular intervals, for specific durations, or targeting certain pages. These heuristics are intentionally broad to ensure high precision in bot detection.

\begin{description}
\item[Forged User Agent:] Some user-agents, like those from automation libraries (e.g., python-request), indicate bot activity. Bot operators often disguise them as regular browser user-agents like Firefox. If the user-agent's capabilities do not match a genuine agent, it is likely fraudulent. Human users typically do not modify their user-agents in this manner It is generally assumed that human users do not alter their user-agents in this way \cite{forged-user-agents-1,forged-user-agents-2}.

\item[Time Between Hits Similarity:] The interval between website visits can indicate bot activity \cite{time-between-hits-1,time-between-hits-2}. Consistent timing between requests suggests automated browsing, as human interactions typically exhibit variability due to innate randomness. For example, a monitoring bot might be programmed to visit a website hourly with precise intervals. 

\item[Unrealistic Window Sizes:] Effective human web browsing requires sufficiently large browser windows for clear graphical display. Bots, conversely, often utilize minimal window sizes, like those used by deep learning bots employing convolutional neural networks \cite{screen-resolution-1}. Smaller windows demand less processing power, which is cost-effective. Therefore, very small browser dimensions, such as an axis under 50 pixels, strongly suggest bot traffic.
\end{description}

\subsection{Bot Detection Using Technical Features}
This approach employs a SGAN trained on a mix of labeled and unlabeled data to analyze individual hits for bot-related characteristics. The choice of an SGAN is strategic due to its proficiency in processing both labeled and unlabeled data. Within this framework, a classifier is trained using the labeled data to differentiate between bots and humans. The SGAN's generator and discriminator are trained using both labeled and unlabeled data; the generator aims to create realistic labeled data points, while the discriminator's objective is to identify fabricated data points. Any generated data points not recognized as artificial are further utilized to refine the classifier. This approach enables the exploitation of features from unlabeled data in classifier training, making the SGAN model particularly suitable for addressing web bot challenges. Web log datasets, like the one used in this research, often contain a significant portion of unlabeled data points.
The labels for training are derived from the assumptions regarding bot and human clients outlined in Section \ref{susec:data-labeling-approach}. These labels allow for confident categorization of the behavioral aspects of each data point, which are then employed to train the binary classifier within the SGAN's discriminator.

The primary goal of adversarial learning is to refine the generator to produce synthetic samples indistinguishable from real ones by the discriminator. However, in the context of bot detection, where the focus is on identifying rather than generating realistic bot data, it is necessary to modify the model architecture accordingly. Consequently, the discriminator in our model consists of two components sharing a common neural network framework: the classifier and the discriminator components. The discriminator component fulfills the traditional role of differentiating between fake and real samples. It outputs a probability $p_{real} \in [0,1]$, indicating the likelihood of a sample being real; values near one suggest a real sample, while values close to zero imply a generated sample. The classifier component, meanwhile, is tasked with determining the class membership of a data point by predicting a vector of class membership probabilities $[p_1, ..., p_n]$ for $n$ classes, utilizing the Softmax activation function. Despite their distinct objectives, the integration of shared weights enhances the classifier's performance by allowing it to benefit from both supervised and unsupervised learning through the discriminator. This is particularly advantageous when labeled data is scarce.

During the training phase, the classifier is trained with supervised samples. Its objective is to predict a vector of class membership probabilities for each sample and compute a loss value $L_C$ using the cross entropy loss function \cite{crossentropy-softmax}. The cross entropy loss function, represented as $L$, is generally defined in equation 4.1. Here, $n$ represents the number of classes, and $k$ refers to the k-th class. Additionally, $Y_k$ indicates the true class value, being one if a sample belongs to class $k$ and zero otherwise. $p_k$ denotes the class membership probability resulting from the Softmax-activated output layer. The loss is minimized when the predicted probabilities align closely with their respective true values.

\begin{equation}
	L(Y_k, p_k) = (-1) \cdot \sum^{n}_{k=1}Y_k \cdot log(p_k)
\end{equation}

Both the discriminator and classifier components in our model utilize the cross entropy loss function, leading to their respective loss values being represented as $L_D$ for the discriminator and $L_C$ for the classifier. The class membership probabilities essential to this process are calculated by applying the Softmax activation function to the output layer of the shared neural network \cite{crossentropy-softmax,keras-sgan}. The Softmax function, denoted as $S$, is defined in equation 4.2. Here, $Z$ represents the vector of all output activations, and $z_i$ signifies the activation of the i-th neuron in the output layer. Softmax is responsible for converting these real-valued inputs into normalized probabilities. The output is a categorical probability distribution, indicating the likelihood of each class membership.

\begin{equation}
	S(Z, z_i) = \frac{e^{z_i}}{\sum_{z_k \in Z}e^{z_k}}
\end{equation}

The loss value is critical for optimizing the weights of the shared neural network through the backpropagation algorithm \cite{backprop}. The discriminator, trained on both real and synthetic samples, aims to enhance its proficiency in distinguishing between these two types. The loss value $L_D$, derived from the discriminator's predictions, is backpropagated through both the discriminator and generator networks to refine the generator's performance. As the generator produces labeled, realistic data points, it significantly augments the classifier's efficacy by providing a larger pool of labeled data.

Distinct from the classifier, the discriminator employs a unique method for its output layer activation, a technique developed by Salimans et al. to optimize the training of generative adversarial networks \cite{exp-sum-activation}. This method, termed the \textit{ExpSum Activation Function} and denoted as $E$, is formulated in equation 3. Here, $Z$ represents the $k$ activations in the output layer of the shared neural network, each playing a role in the improved discrimination of real versus fake data points.

\begin{equation}
E(Z) = \frac{F(Z)}{F(Z)+1} \qquad F(Z) = \sum_{z_k \in Z} e^{z_k}
\end{equation}

\subsection{Bot Detection through Analysis of Website Traversal Graphs}
The appealing aspect of these graphs is that they are purely based on behavioral features. Website traversal graphs, which depict user navigation on a website with nodes representing sub-pages and edges as navigational links, are further enhanced by weighting edges according to visit frequency and labeling nodes with pertinent attributes. This enriched graphical representation serves as a comprehensive data source for analysis. Our core hypothesis posits that automated web bots exhibit unique navigation patterns, distinct from those of human users, characterized by intensive search tactics and specific page refresh frequencies.

Our approach involves a three-layered methodology. The first layer involves extracting prominent features from the raw data. Following this, a sorting pooling layer restructures these features into a format conducive to deep learning analysis. The final stage employs a one-dimensional Convolutional Neural Network (CNN). Notably, incoming data points, termed as hits, incrementally expand the existing session graph, allowing for dynamic graph development. The classification process thus takes into account the entirety of session hits, rather than isolated events.

Table \ref{tab:class-properties} outlines the specific features utilized in our analysis and their association with the elements of the Website Traversal (WT) graph.

\begin{table*}[!htbp]
\centering

\setlength{\tabcolsep}{10pt} 

\begin{tabular}{l p{4.5cm} l}
\hline
\textbf{Attribute} & \textbf{Description} & \textbf{Component} \\
\hline
First Hit Pagename & The initial page of a session & Node \\
Detailed Pagename & The specific page related to a hit & Node, Edge \\
Previous Pagename & The preceding page in a hit & Node, Edge \\
Timestamp & The time of page visit & Node Label \\
Page Type & The category of the visited page & Node Label \\
Benchmark Label & Associated with a specific hit's benchmark label & Node Label \\
\hline
\end{tabular}
\caption{Features Incorporated in WT Graphs}
\label{tab:class-properties}
\end{table*}

Furthermore, the following metrics are extracted from WT graphs:
\begin{description}[itemsep=0pt]
\item [Node Degree] The count of edges linked to a node.
\item [Node Count] The aggregate number of nodes within the graph.
\item [Edge Count] The complete count of edges in the graph.
\item [Page Type Distribution] This metric calculates the relative frequency of various page types within a set of hits, based on their occurrence and total number of hits for each page type in the set, applicable to a single session or WT Graph.
\item [Session Topics] Extracted using the RAKE algorithm \cite{rake}, this metric comprises keywords from page titles within the WT graph, focusing only on those with a score of 1 or higher. It aims to differentiate between known bots and humans by comparing session topic sets of unidentified clients.
\item [Number of Hits] The total visits across all pages in the graph.
\item [Hits per Sub Page] The visit count for each subpage.
\item [Degree Centrality] A metric indicating a node's centrality in the graph, based on its connected edges.
\item [Betweenness Centrality] This measure reflects a node's significance in the graph, determined by the number of shortest paths passing through it among all other nodes.
\end{description}

\section{Implementation}
\subsection{SGAN Architecture and Training}
The SGAN used in the evaluation comprises a discriminator and a generator network. The discriminator integrates parallel, aligned layers for discrimination and classification. It employs binary cross-entropy for the discriminator layer and sparse categorical cross-entropy, coupled with a Softmax activation function, for the classifier layer. Both layers utilize the Adam optimization algorithm, with learning rate parameters set at $\alpha = 0.0002$ and $\beta_1 = 0.5$. The architecture features seven shared hidden layers: the first, third, and fifth are dense layers with 100 units each, activated by the Sigmoid function. The second, fourth, and sixth layers are leaky ReLU layers with a fixed scaling parameter of $\alpha = 0.2$. A dropout layer with $p = 0.4$ constitutes the seventh layer. The generator network comprises three layers: an input layer drawing from a 100-dimensional latent vector with 200 units activated by the Sigmoid function, and an output layer with units corresponding to each dataset feature, activated by ReLU. The SGAN employs binary cross-entropy for loss and Adam for optimization, mirroring the discriminator's initialization.

\subsection{DGCNN Architecture and Training}
For implementing bot detection via WT graphs, we follow the framework presented in the study by \cite{dgcnn}. This research introduces a graph-based deep learning mechanism for classification tasks, enabling the encoding of undirected graphs with node features for neural network training.

The DGCNN consists of a GCN and a 1D-CNN. The GCN section includes four graph convolution layers with 32, 32, 32, and 1 hidden units, respectively, followed by a sort pooling layer to format outputs for the 1D-CNN. Activation in each layer is achieved through $tanh$. The GCN parameter $k$ is fixed at 35. The 1D-CNN features two convolutional layers: the first with 16 filters, kernel size, and stride matching the sum of hidden units in all GCN layers; the second with 32 filters, a kernel size of five, and a stride of one. A max pooling layer with a pool size of two separates these layers. Following the second convolutional layer is a dense layer with 128 hidden units activated by ReLU, succeeded by a dropout layer with $p = 0.5$. The final output layer, activated by Sigmoid, uses a single unit. Binary cross-entropy serves as the loss function, and Adam, parameterized with a learning rate of $\alpha = 0.0001$, is the optimization algorithm.

\section{Evaluation}
\label{sec:eval}
The evaluation of the proposed methodologies encompasses both individual and collective assessments. Specifically, this paper aims to elucidate the following research inquiries:

\begin{description}
\item[RQ1:] What is the efficacy of each module within the detection pipeline in differentiating between bots and human users?
\item[RQ2:] Which attributes exert the most significant impact on the outcomes of detection?
\item[RQ3:] How does the dimensionality of a WT graph influence the classification accuracy?
\end{description}

\subsection{Data Labeling and Ground Truth}
\label{susec:data-labeling-approach}
Our analysis leverages a dataset from an e-commerce website, which garners approximately 40 million monthly visits. The evaluation is conducted on a subset of around 1.4 million of these visits. To our knowledge, there currently exists no standardized dataset suitable for this type of evaluation.

There are several advantages to utilizing a real-world dataset, such as its realism and the inclusion of unknown types of bots. However, a significant drawback is the lack of ground truth, specifically identifying who is a bot and who is not. We address this issue by using assumptions to provide labeling of the data. 

\begin{description}
\item [Human Assumption] Traffic is labeled as human if it originates from accounts of human users (i.e., not automated test users) within the organization hosting the website. This is based on the assumption that it is highly unlikely for an employee to create a bot to interact with their own company's website, as this would risk potential employment termination for engaging in malicious activities.
\item [Bot Assumption] A client is categorized as a bot if their requests originate from an IP address associated with a cloud provider's computational center. While it is acknowledged that some legitimate users may access websites via proxies or VPNs, this is considered a minority scenario.
\end{description}

It is important to note that generating ground truth through assumptions is a practice that has been used previously. However, we adopt a stricter approach than related work. For instance, Botcha's \cite{botcha} hypothesis that bots do not engage in purchases is increasingly being questioned. Modern examples demonstrate that bots actively purchase items such as graphics cards, concert tickets, and game consoles \cite{ecommerce-scalper,ticketmaster-scalper}.

Table \ref{tab:ground-truth} illustrates the quantity of data points labeled under the human and bot assumptions, in conjunction with those identified by the heuristics. It is critical to acknowledge, however, that the heuristics—excluding the human assumption—are incapable of discerning human users.
Additionally, our analysis examined potential conflicts between the heuristics and our fundamental assumptions. We discovered that the heuristics erroneously classified 9 instances of human interactions as bot activities. Despite this, the obtained recall rate of 0.9988 was deemed satisfactory for our purposes.

\begin{table}
\centering
\setlength{\tabcolsep}{5pt} 
\begin{tabular}{l c c}
\textbf{Class} & \textbf{Assumption} & \textbf{Enhanced by} \\
& \textbf{(\#Hits)}   & \textbf{Heuristics (\#Hits)} \\
\hline
Bot & 51.462 & 65.018 \\
Human & 7.630 & 7.630\\
Unknown & 723.579 & 710.023\\
\hline
\end{tabular}
\caption{Ground Truth Based on Initial Assumptions and Heuristic Refinement}\label{tab:ground-truth}
\end{table}

Table \ref{tab:ground-truth} presents categorizes individual requests from our dataset based on their identification as either human, bot, or unknown. It is structured to compare the number of hits identified through our initial assumptions (cf. \ref{susec:data-labeling-approach}) with those further refined by the application of our heuristics (cf. \ref{sususec:heuristic-bot-detection}), which are designed to identify more bots but not humans. It shows a refinement in bot detection in our dataset, increasing from 51,462 to 65,018 instances with the application of heuristics, while maintaining consistent human identification and reducing unknown classifications.

\subsection{Results}
In this section, we present the results of our comparative analysis.

\subsubsection{RQ1: Detection Performance}

We compare the performance of the model relying on technical features like window size (SGAN) and the model relying on behavior only (DGCNN) with another bot detection approach on the same dataset. Botcha was configured with the exact parameters described in its publication \cite{botcha}.

\begin{table*}
\centering
\setlength{\tabcolsep}{5pt}
\begin{tabular}{l c c c c c}
\textbf{Model} & \textbf{Accuracy} & \textbf{Recall} & \textbf{Precision} & \textbf{F1-Score} & \textbf{AUROC} \\
\hline
SGAN & 0.9895 & 0.9875 & 0.9189 & 0.9519 & 0.9886\\
DGCNN & 0.9845 & 0.9833 & 0.9791 & 0.9812 & 0.9892\\
\hline
Botcha-MAM & 0.9364 & 0.8383 & 1.0 & 0.9120 & 0.9437\\
Botcha-RAM & 0.9952 & 0.9663 & 0.9807 & 0.9735 & 0.9996\\
\hline
\end{tabular}
\caption{Comparison to Related Work (Botcha)}
\label{tab
}
\end{table*}

All approaches exhibit strong performance. SGAN and Botcha, which rely on technical (non-behavioral) attributes, might seem more effective. However, the superiority of WT graphs that leverage behavioral features is noteworthy. This aspect is increasingly significant in the dynamic and challenging landscape of bot detection, where bot creators constantly adapt to evade detection mechanisms.

Behavioral features are paramount due to their independence from the underlying automation technology. For instance, while using a real, rendered browser enables bots to mimic human activity and evade detection based on technical, non-behavioral features, achieving this level of mimicry requires sophisticated emulation of human behavior. Bot creators encounter difficulties in replicating the nuanced user behavior unique to each website, which varies with the site's type and structure. Furthermore, programming bots to mimic human-like browsing can significantly hinder their efficiency. For example, adhering to human browsing patterns, such as generating an average of 10 results per minute over 5-minute sessions, compromises the inherent speed and endurance advantages of bots.

In this landscape, while SGAN and DGCNN demonstrate commendable performance, with SGAN slightly ahead in terms of accuracy and AUROC, Botcha-RAM stands out as the most effective model, achieving high scores in accuracy, recall, precision, F1-Score, and AUROC. Botcha-MAM, despite its unmatched precision, shows reduced efficiency in other essential metrics.

\subsubsection{RQ2: Technical Feature Importance}
In our investigation into SGAN's detection capabilities, we aim to identify which features exert the most significant impact. To this end, we employ the Permutation Importance Algorithm \cite{permutation-importance}, a method effective in discerning key features influencing data point classification. Initially, the algorithm calculates a reference score \( s \), assessing the classifier's performance on a particular dataset, using a chosen scoring function. 

The process involves iterating over each feature column \( d = 1, \ldots, D \), where \( D = |\textit{features}| \). For every column, the algorithm shuffles the data \( K \) times, with \( k = 1, \ldots, K \), and computes a new score \( s_{k,d} \) for the perturbed dataset. It then evaluates how this new score compares with the reference score \( s \). This comparison is crucial as it reveals each feature's effect on classification accuracy. A significant drop in the score due to feature manipulation signals a high importance of that feature, while a negligible change indicates low importance.
The importance of a feature is quantified using the equation:  \(i_d = s - \frac{1}{K} \sum_{k=1}^{K} s_{k,d}\)

Applying the SGAN classifier to the test dataset, we determine the feature importances, as detailed subsequently. Notably, the algorithm was configured with a \( K \) value of 50, and various scoring functions were employed. Additionally, the application of Onehot encoding to some features accounts for the presence of specific values in feature names.

\begin{table}
\centering
\begin{tabular}[c]{c| c c c c c c c}
    \hline
    \textbf{Feature} & \multicolumn{2}{c}{\textbf{R2-Score}} & \multicolumn{2}{c}{\textbf{Negative MSE}}\\
    \cline{2-5}
     & $\mu_{i}$ & $\sigma_{i}$ & $\mu_{i}$ & $\sigma_{i}$ \\
     \hline
    post\_browser\_height & 0.542  & $\pm$ 0.008 & 0.051 & $\pm$ 0.001 \\
    post\_browser\_width & 0.287 & $\pm$ 0.010 & 0.027 & $\pm$ 0.001 \\
    post\_java\_enabled\_N & 0.082 & $\pm$ 0.003 & 0.008 & $\pm$ 0\\
	post\_java\_enabled\_Y & 0.061 & $\pm$ 0.002 & 0.006 & $\pm$ 0\\
	user\_agent\_Other & 0.024 & $\pm$ 0.002 & 0.002 & $\pm$ 0\\
	visit\_page\_num & 0.022 & $\pm$ 0.003 & 0.002 & $\pm$ 0\\
	visit\_num & 0.012 & $\pm$ 0.004 & 0.001 & $\pm$ 0\\
	hourly\_visitor & 0.010 & $\pm$ 0.001 & 0.001 & $\pm$ 0\\
	page\_type\_product & 0.005 & $\pm$ 0.001 & 0 & $\pm$ 0\\
	last\_purchase\_num & 0.004 & $\pm$ 0.001 & 0 & $\pm$ 0\\
	user\_agent\_Mozilla/5.0 & 0.003 & $\pm$ 0.001 & 0 & $\pm$ 0\\
 \hline

\end{tabular}
\caption{Feature Importance Scores of the SGAN Classifier for the most important features. MSE: Mean Squared Error. $\mu$: Mean accuracy decrease. $\sigma$: Standard deviation.}
\label{tab:feature_importance}
\end{table}

As shown in Table \ref{tab:feature_importance}, the most prominent features in bot detection are attributes such as \textit{post\_browser\_height} and \textit{post\_browser\_width}. These features, despite their high significance as shown by their R2-scores and Negative MSE values, are relatively easy for bots to falsify. By simply adjusting the browser height and width to mimic those of a typical user, bots can effectively camouflage their non-human nature.
This insight highlights the imperative of integrating behavioral characteristics into bot detection mechanisms. Behavioral features explore the intricacies and patterns inherent in human interactions, which are considerably more arduous for bots to emulate. Contrasting with static attributes such as browser dimensions or Java enablement status, behavioral patterns encompass intricate, dynamic, and frequently nuanced human activities. Even if a bot were to mimic these patterns, it would result in a significant loss of efficiency.

\subsubsection{RQ3: WT Graph Size Importance}
For this investigation, we utilized a dataset formed through overlap clustering, assigning 30 percent as test data. The most efficacious DGCNN model, pretrained on 70\% of the dataset and employing overlap clustering, was used in this study. It is important to note that the AUROC metric was excluded from use, given its inapplicability in certain graph sizes where only a single class exists.

\begin{table}
\centering
\setlength{\tabcolsep}{5pt} 
\begin{tabular}[c]{c c c c c c c c}
\hline
\textbf{Nodes} & \textbf{\# Graphs} & \textbf{ACC} & \textbf{Recall} & \textbf{Precision} & \textbf{F1-Score} \\
\hline
    1 & 26137 & 0.998 & 0.981 & 0.998 & 0.99\\
    2 & 17066 & 0.973 & 1.0 & 0.974 & 0.986\\
    3 & 3533 & 1.0 & 1.0 & 1.0 & 1.0\\
    4 & 371 & 0.998 & 0.999 & 0.999 & 0.999\\
    5 & 251 & 0.998 & 1.0 & 0.998 & 0.999\\
    6 & 101 & 0.998 & 1.0 & 0.998 & 0.999\\
    7 & 526 & 0.997 & 1.0 & 0.997 & 0.998\\
    8 & 1579 & 1.0 & 1.0 & 1.0 & 1.0\\
    9 & 1175 & 1.0 & 1.0 & 1.0 & 1.0\\
    10 & 108 & 1.0 & 1.0 & 1.0 & 1.0\\
    \hline
\end{tabular}
\caption{Classification performance depending on WT graph size.}
\label{tab:wt_graph_size_results}
\end{table}

The findings indicate that an increase in graph size correlates with enhanced performance of the model. The data presented in the table reveals that even minimal graphs, comprising one to three nodes, are classified with high accuracy. This underscores the efficacy of WT graphs in encapsulating web client characteristics, thereby facilitating the DGCNN's ability to learn representations, leading to heightened accuracy, precision, and recall.
It is noteworthy that this performance is attainable due to the unique feature of WT graphs, which aggregate multiple interactions with the same webpage into a singular node. This characteristic enables the trained DGCNN to identify bots effectively, even in scenarios where they interact with only a single sub-page multiple times. Additionally, the classification's performance augments with the expansion of nodes in the WT graph, suggesting that the DGCNN's bot detection capabilities are more pronounced as clients engage with a greater number of sub-pages.

\subsection{Limitations}
There are several limitations to this paper that should be considered.
First, we are unable to share the dataset that was used for this research due to data protection reasons. This limitation is also present in many related works, and thus limits our ability to fully benchmark our approach against these studies. 

Since we are using a real world dataset we lack an accurate ground truth. We rely instead of the most basic assumptions we can think of~(cf. \ref{susec:data-labeling-approach}). However, while we have considered the characteristics and behaviors of both groups in the design of our approach, it is possible that some bots or humans may exhibit behaviors that we have not accounted for. This could potentially impact the accuracy of our method in detecting certain types of bots.

Additionally, it is worth noting that our approach may not be able to detect bots that behave exactly like humans. This however is a common limitation among bot detection approaches. However, a bot that behaves exactly like a human is less effective for the bot operator and makes it more difficult to perform malicious activities.

\section{Conclusion}
\label{sec:conclusion}
In conclusion, our bot detection framework, applied to a large-scale e-commerce site with a substantial monthly visitor count of 50 million, demonstrates its robustness and effectiveness. The methodology begins with heuristic-based approaches for simple bot detection, and then advances to more sophisticated techniques involving technical features analyzed through a Semi-supervised Generative Adversarial Network (SGAN). Behavioral features, particularly those related to website navigation patterns transformed into a Website Traversal Graph and processed via a Deep Graph Convolutional Neural Network (DGCNN), further strengthen our detection capabilities.

Our assessment, when benchmarked against the Botcha methodology, demonstrates that our approach achieves a level of effectiveness similar to theirs, albeit solely relying on behavioral features. The focus on behavioral analysis not only yields high detection accuracy but also necessitates a greater level of sophistication from bot developers. This requirement for bots to more accurately imitate human behavior leads to a decrease in their effectiveness and operational viability, thus enhancing the efficacy of our bot detection framework.

Future work will explore the development of more reliable measures that require less user interaction up to the point of detection, aiming to enhance the efficiency in identifying bots.
\newpage

\bibliographystyle{plain}

%\bibliography{esorics-secai/bib}

\appendix

\end{document}